\documentclass[journal]{IEEEtran}

\usepackage{cite}
\usepackage{graphicx}
\usepackage{multirow}
\usepackage[pagebackref=true,breaklinks=true,colorlinks,bookmarks=false]{hyperref}
\usepackage{amsmath}
\usepackage{amssymb}
\usepackage{color}

\usepackage{pgfplots}
\pgfplotsset{width=9cm, compat=1.9}

\newcommand{\etal}{\textit{et al.}\@ }
\newcommand{\ie}{\textit{i.e.}\@ }

\begin{document}

\title{Rethinking Image Skip Connections in StyleGAN2}

\author{Seung Park and Yong-Goo~Shin,~\IEEEmembership{Member}, IEEE

\thanks{S. Park is with Biomedical Engineering, Chungbuk National University Hospital, Chungbuk National University College of Medicine, Cheongju-si, Chungcheongbuk-do, 28644, Rep. of Korea (e-mail: spark.cbnuh@gmail.com, spark90@chungbuk.ac.kr). }

\thanks{Y.-G. Shin is with the Department of Electronics and Information Engineering, Korea University, Sejong-si, Rep. of Korea (e-mail: ygshin92@korea.ac.kr, corresponding author)}

\thanks{This work was supported by the National Research Foundation of Korea (NRF) grant funded by the Korea government (MSIT) (No.2022R1G1A1004001). This research was supported by the MSIT (Ministry of Science and ICT), Korea under the ITRC(Information Technology Research Center) support program (IITP-2023-RS-2023-00258971) supervised by the IITP (Institute for Information \& Communications Technology Planning \& Evaluation).}

}


\markboth{Submitted to IEEE Transactions on Neural Networks and Learning Systems}
{Shell \MakeLowercase{\textit{Shin et al.}}}

\maketitle

\begin{abstract}
Various models based on StyleGAN have gained significant traction in the field of image synthesis, attributed to their robust training stability and superior performances. Within the StyleGAN framework, the adoption of image skip connection is favored over the traditional residual connection. However, this preference is just based on empirical observations; there has not been any in-depth mathematical analysis on it yet. To rectify this situation, this brief aims to elucidate the mathematical meaning of the image skip connection and introduce a groundbreaking methodology, termed the image squeeze connection, which significantly improves the quality of image synthesis. Specifically, we analyze the image skip connection technique to reveal its problem and introduce the proposed method which not only effectively boosts the GAN performance but also reduces the required number of network parameters. Extensive experiments on various datasets demonstrate that the proposed method consistently enhances the performance of state-of-the-art models based on StyleGAN. We believe that our findings represent a vital advancement in the field of image synthesis, suggesting a novel direction for future research and applications.

\begin{IEEEkeywords}
Generative adversarial network, image synthesis, StyleGAN, image skip connection
\end{IEEEkeywords}

\end{abstract}

\section{Introduction}
\label{sec1}
\IEEEPARstart{G}{enerative} adversarial networks (GANs)~\cite{goodfellow2014generative} have shown impressive outcomes in various computer vision tasks including image synthesis~\cite{lee2022generator, bai2023glead, park2024novel, yeo2021simple, sagong2022conditional}, image-to-image translation~\cite{isola2017image, choi2018stargan, zhu2017unpaired}, and image inpainting~\cite{sagong2019pepsi, shin2020pepsi++}. In general, the GANs framework comprises two distinct networks: the generator (G) and the discriminator (D). G is tasked with producing data that mimics the distribution of real data, aiming to deceive D. Conversely, D pours attention on distinguishing the synthesized samples from the real ones. Ideally, the process would culminate in an optimal solution where G can perfectly mimic the real data distribution, making it nearly impossible for D to discern the source of the images~\cite{goodfellow2014generative}.

However, the competition between G and D is not fully balanced. Specifically, D, by evaluating the difference between real and synthesized samples, assumes a dual role in this adversarial framework: as both competitor and arbiter. This configuration allows D to dominate the game and becomes a primary cause of instability of GANs training. To alleviate this problem, some papers~\cite{miyato2018spectral, gulrajani2017improved, zhang2019consistency, arjovsky2017wasserstein, wu2019generalization, wei2018improving, kurach2019large} introduced regularization or normalization techniques that inhibit the discriminator from making the sharp gradient space. In traditional regularization approaches~\cite{gulrajani2017improved, wu2019generalization, wei2018improving}, gradient penalty-based methods have become particularly popular. These techniques, which integrate the gradient norm as a penalty term into the adversarial loss function, have been extensively adopted across various applications for their effectiveness. Normalization techniques~\cite{miyato2018spectral, arjovsky2017wasserstein, kurach2019large} have also received significant attention in research. Among these, spectral normalization (SN) emerges as the most notably recognized method. Since the SN method ensures adherence to the Lipschitz constraint for discriminators by regulating the SN of each layer, it plays a crucial role in stabilizing training processes. These normalization and regularization methods have been extensively adopted owing to their critical role in improving the stability of GAN training.

On the other hand, explorations into architectural enhancements for GANs have led to significant developments in performance as well as training stability~\cite{park2021GRB, miyato2018cgans, park2021generative_conv, zhang2019self, yeo2021simple}. Miyato~\etal~\cite{miyato2018cgans} proposed a novel conditional projection approach, a technique that integrates conditional information into D directly. This approach has significantly improved the performance of conditional GANs (cGANs) through a straightforward method. Zhang~\etal~\cite{zhang2019self} proposed a novel self-attention module that guides G and D on where to focus, whereas Yeo~\etal~\cite{yeo2021simple} introduced a cascading rejection module that iteratively computes the adversarial loss using varied feature vectors. Li~\etal~\cite{li2023dw} proposed a dynamic weighted GAN, called DW-GAN, which synthesizes the images following the color tone of the target image. These techniques offer the advantage of significantly enhancing the image synthesis performance of GANs through simple methods.

Recently, StyleGAN~\cite{karras2019style} has emerged as the baseline model in the GAN-based image synthesis domain.  This method is celebrated for its ability to produce images of exceptional quality and for effectively mitigating the issue of training instability. Unlike with the previous studies~\cite{miyato2018spectral, gulrajani2017improved, zhang2019consistency, arjovsky2017wasserstein, wu2019generalization, wei2018improving, kurach2019large}, StyleGAN allows G to control the synthesis process of the image at different levels of detail through the use of \textit{Style} parameters. In detail, \textit{Style} parameters are extracted from a latent space and serve to modulate the mean and variance within the framework of adaptive instance normalization (AdaIN)~\cite{huang2017arbitrary}, enabling control over characteristics of the generated images. To more effectively control \textit{Style} parameters, a variety of models such as StyleGAN2~\cite{karras2020analyzing}, StyleGAN3~\cite{karras2021alias}, and StyleGAN-XL~\cite{sauer2022stylegan} have been developed, where these technologies are capable of producing images with the high quality.

Among the StyleGAN variations, StyleGAN2~\cite{karras2020analyzing} is being widely adopted as the baseline model in recent research papers~\cite{lee2022generator, bai2023glead, park2024novel, sauer2023stylegan, wang2023styleinv, yildirim2023diverse, zhang2023getavatar}. In~\cite{karras2020analyzing}, they conducted a variety of studies, including innovative ways to utilize \textit{Style} parameters, new regularization techniques, and an analysis of issues with the progressive growing technique. Among these diverse research efforts, we have delved into the significance of using image skip connections as an alternative to the progressive growing technique. In~\cite{karras2020analyzing}, they have conducted experiments to evaluate performance by applying residual connections and image skip connections to G instead of using the progressive growing technique. Through numerous experiments, they proved that applying the image skip connection to G is more effective than the residual connection. However, these findings are based on empirical evidence rather than a mathematical analysis of the image skip connection; the conclusions drawn do not stem from a theoretical examination of their underlying principles.

To the best of our knowledge, however, there has been no prior research that has pointed out this issue. To address this issue, this brief first analyzes the mathematical meaning of the image skip connection and discusses the reasons for their superior performances compared to the residual connection. Then, we introduce a novel skip connection approach, called the image squeeze connection, which consistently surpasses the performance of the image skip connection in cutting-edge models derived from StyleGAN2. Furthermore, we will show that the proposed method achieves superior performance despite necessitating fewer network parameters compared to the original G in StyleGAN2. To validate the superiority of its proposed method, this brief presents comprehensive experimental findings across a variety of datasets such as CIFAR-10~\cite{krizhevsky2009learning}, FFHQ~\cite{karras2019style}, subsets of LSUN~\cite{yu15lsun}, and AFHQ~\cite{choi2020stargan}. Moreover, a series of ablation studies were conducted to highlight the robust generalization capabilities of the proposed method. Through quantitative evaluations, it is clear that the proposed method not only significantly boosts GAN performance but also achieves remarkable enhancements in key metrics, including the Inception score (IS)~\cite{salimans2016improved}, Frechet Inception Distance (FID)~\cite{heusel2017gans}, and Precision/Recall~\cite{kynkaanniemi2019improved, sajjadi2018assessing}. The proposed method can be seamlessly integrated into any setup, as it effectively reduces the number of network parameters by approximately 12.1\%. Our contribution can be summarized as follows:

\begin{itemize}
\item We propose a novel approach, called image squeeze connection, which is simple but consistently improves the performance of various methods based on StyleGAN2.
\item Experimental results on several datasets show that our method works better than previous approaches, achieving higher scores in key performance measures.
\item The proposed method simplifies and streamlines the baseline model by reducing hardware costs by approximately 12.1\%.

\end{itemize}

\section{Preliminaries}
\label{sec2}
\subsection{Generative Adversarial Networks}
\label{sec2.1}
GANs~\cite{goodfellow2014generative} consist of two neural networks, the generator (G) and the discriminator (D), which are trained simultaneously through a competitive process. The role of G is to generate data that is indistinguishable from real data, while the job of D is to distinguish between the fake data synthesized by G and actual data. This adversarial process is akin to a forger trying to create a counterfeit painting, with D acting as the art critic trying to spot the fake. Through this competition, G could produce fake data that looks real theoretically. The traditional objective functions of G and D are defined as follows:

\begin{eqnarray}
\label{eq1}
    L_G = -E_{z\sim P_{z}(z)}[\log(D(G(z)))],
\end{eqnarray}
\begin{eqnarray}
\label{eq2}    
      \lefteqn{L_D = - E_{x\sim P_\textrm{data}(x)}[\log D(x)]}\nonumber\\
    & & {\qquad \qquad} - E_{z\sim P_{z}(z)}[\log(1-D(G(z)))]. 
\end{eqnarray}

However, a notable challenge inherent in the original objective functions is the instability during the training phase. This instability often manifests as difficulty in convergence, leading to sub-optimal generation of data and sometimes causing the training process to fully fail. To address these issues, many papers~\cite{mao2017least, gulrajani2017improved, arjovsky2017wasserstein, mescheder2018training, karras2020analyzing} have been actively proposing and evaluating alternative objective functions that can offer more stable training dynamics.

In the current landscape of GAN research and application, a particularly effective approach has gained prominence, known as the non-saturating logistic loss with $R_{1}$ regularization~\cite{mescheder2018training, karras2020analyzing, karras2019style, karras2021alias}. This method enhances the training stability and the quality of the generated images by carefully balancing the learning process of G and D. The non-saturating logistic loss ensures that the gradients of G do not vanish or explode, maintaining a steady pressure on the generator to improve. Concurrently, the $R_{1}$ regularization adds a penalty based on the first-order gradient of the D for real data, encouraging smoother decision boundaries and further stabilizing the training dynamics. The non-saturating logistic loss with $R_{1}$ regularization is defined as below:

\begin{eqnarray}
\label{eq3}
      \lefteqn{L_D = E_{x\sim P_\textrm{data}(x)}[\log(1 + e^{-D(x)}) + \cfrac{\gamma}{2}\|\nabla D(x)\|^{2} ]}\nonumber\\
    & & {\qquad \qquad} + E_{z\sim P_{z}(z)}[\log(1 + e^{D(G(z))})],
\end{eqnarray}
\begin{eqnarray}
\label{eq4}
    L_G = E_{z\sim P_{z}(z)}[\log(1 + e^{-D(G(z))})],
\end{eqnarray}
where $\gamma$ is a hyper-parameter that constrains the gradient magnitude of the discriminator. By using this function, many GAN-based image synthesis methods have achieved more reliable and higher-quality results.

\subsection{Revisit StyleGAN2}
\label{sec2.2}
StyleGAN2~\cite{karras2020analyzing} is an advanced version of the original StyleGAN architecture, which represents a significant advancement in the field of GANs. StyleGAN2 introduces several key improvements over its predecessor to enhance image quality and address specific issues observed in StyleGAN-generated images, such as unnatural artifacts and inconsistencies. One of the notable changes in StyleGAN2 is the redesign of the normalization method that maintains stylistic control while reducing artifacts. In detail, the original StyleGAN used a feature called AdaIN (Adaptive Instance Normalization) at each layer of G to inject style information. However, this led to certain artifacts and issues related to style control. StyleGAN2 introduces a revised approach to normalization, called weight demodulation. This technique replaces the previous modulation mechanism and ensures that the scaling of the convolutional weights does not introduce undesired artifacts in the generated images. Weight demodulation helps in producing cleaner and more coherent textures, significantly improving image quality.

Another improvement is in the handling of the progressive growing technique in StyleGAN. Instead of progressively increasing the resolution during training, StyleGAN2 modifies the network architecture and training process to maintain a fixed resolution, which helps reduce training instability and eliminates phase artifacts. Specifically, they conducted extensive experiments to evaluate the effectiveness of different strategies that improve the information flow within the network, such as residual connections~\cite{he2016deep, miyato2018spectral, gulrajani2017improved}, image skip connections~\cite{karnewar2020msg, ronneberger2015u}, and hierarchical methods~\cite{zhang2017stackgan, zhang2018stackgan++}. Based on the experimental results, they concluded that integrating G with image skip connections and D with residual connections leads to effective performance. It is reliable to draw conclusions based on the results of various experiments, but we believe that it is necessary to consider the theoretical reasons why image skip connections yield good results. Therefore, in this brief, we first analyze the mathematical meaning of image skip connections and propose a new image skip connection method that shows impressive performance. 

\section{Proposed Method}
\label{sec3}
\subsection{Analysis of the image skip connection}
\label{sec3.1}
\begin{figure}
\centering
\includegraphics[width=0.75\linewidth]{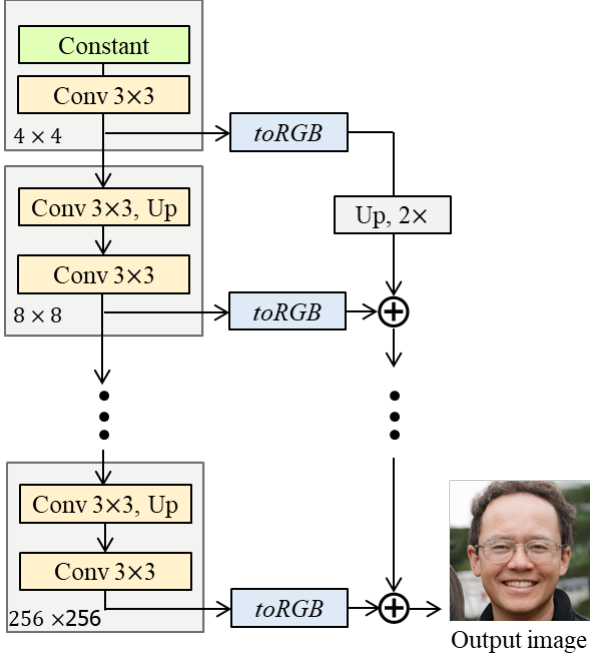}
\caption{The overall architecture of the G in StyleGAN2, which utilizes the image skip connections. The mapping network, which generates the style vector, is excluded for clarity in the illustration.}
\label{fig:fig1}
\end{figure}

As shown in Fig.~\ref{fig:fig1}, StyleGAN2 generates the output image by aggregating the images derived from the intermediate layers. While this appears to be a coarse-to-fine technique where images are created from intermediate layers and then progressively added together, it would not be the case. The reason is as follows: Unlike the convolution layer that conducts both modulation and demodulation (yellow box in Fig.~\ref{fig:fig1}), the \textit{toRGB} layer performs only modulation. In other words, since demodulation is not performed, the operation of the \textit{toRGB} layer is equivalent to first modulating the feature and then performing a $1\times1$ traditional convolution layer without activation function. Therefore, we can express the \textit{toRGB} layer with the following formula:

\begin{equation}
    I_j(x,y) = W_{j}^Tf_{j}'(x,y),
\label{eq6}
\end{equation}
where  $f_{j}' \in \mathbb{R}^{c_j \times h \times w }$ indicates \textit{j-th} intermediate feature modulated by scaling each channel along with the given \textit{Style}, and $I_j \in \mathbb{R}^{3 \times h \times w}$ and $W_{j} \in \mathbb{R}^{c_j \times 3}$ are the \textit{j-th} intermediate image and weight matrix of $1 \times 1$ convolution, respectively. In addition, \textit{x} and \textit{y} represent the pixel coordinates of the feature map, while $c_j$ denotes the channel dimension of $f_{j}'$. Note that there is no non-linearity in Eq.~\ref{eq6} since the modulation and the $1 \times 1$ convolution in each pixel are linear operations. 

Using Eq.~\ref{eq6}, we can express in mathematical terms how the output image is formed. For simplicity, let us assume that the output image has $8\times8$ resolution. Then, the output image $I \in \mathbb{R}^{3 \times 8 \times 8}$ can be formulated as

\begin{equation}
    I(x,y) = Up(I_1(x,y)) + I_2(x,y),
\label{eq7}
\end{equation}
where $Up(\cdot)$ indicates the $2\times$ up-sampling function such as nearest neighbor and bilinear interpolation methods, and $I_1$ and $I_2$ are intermediate images having $4\times4$ and $8\times8$ resolution, respectively. Using Eq.~\ref{eq6}, we can rewrite the Eq.~\ref{eq7} as 

\begin{equation} 
\label{eq8}
\begin{split}
I(x,y) & = Up(I_1(x,y)) + I_2(x,y) \\
 & = Up(W_{1}^Tf_{1}'(x,y)) + W_{2}^Tf_{2}'(x,y) \\ 
 & = W_{1}^TUp(f_{1}'(x,y)) + W_{2}^Tf_{2}'(x,y) \\
 & = W_{a}^Tf_{a}'(x,y), 
\end{split}
\end{equation}
where $W_{a} \in \mathbb{R}^{(c_1 + c_2) \times 3}$ is an integrated matrix formed by the combination of $W_1$ and $W_2$, and $f_{a}' \in \mathbb{R}^{(c_1 + c_2) \times h \times w}$ also indicates an combined feature formed by concatenating $Up(f_{1}')$ and $f_{2}'$ along the channel dimension. That means the existing image skip connection technique achieves mathematically identical results by concatenating the modulated intermediate features, \ie $f_j'$, in the channel direction, followed by performing a 1x1 convolution operation. 


Based on this analysis, we point out that the conventional image skip connection method may encounter problems when generating high-resolution images. For instance, when producing images with a resolution of $256\times256$, image skip connections are executed from a resolution of $4\times4$ up to $256\times256$. As mentioned earlier, the image skip connection technique, which has the same mathematical meaning with concatenating intermediate features at each resolution followed by a 1x1 convolution, leads to an issue where the dimension of the concatenated features increases. In essence, when generating a $256\times256$ image, it is equivalent to projecting modulated features with 2,496 dimensions onto a three-dimensional image using a single linear operation, \ie $1\times1$ convolution layer without actviation function. However, this method faces challenges in achieving optimal performance due to the difficulty of directly projecting high-dimensional features into a lower-dimensional space. 

\begin{figure}
\centering
\includegraphics[width=0.95\linewidth]{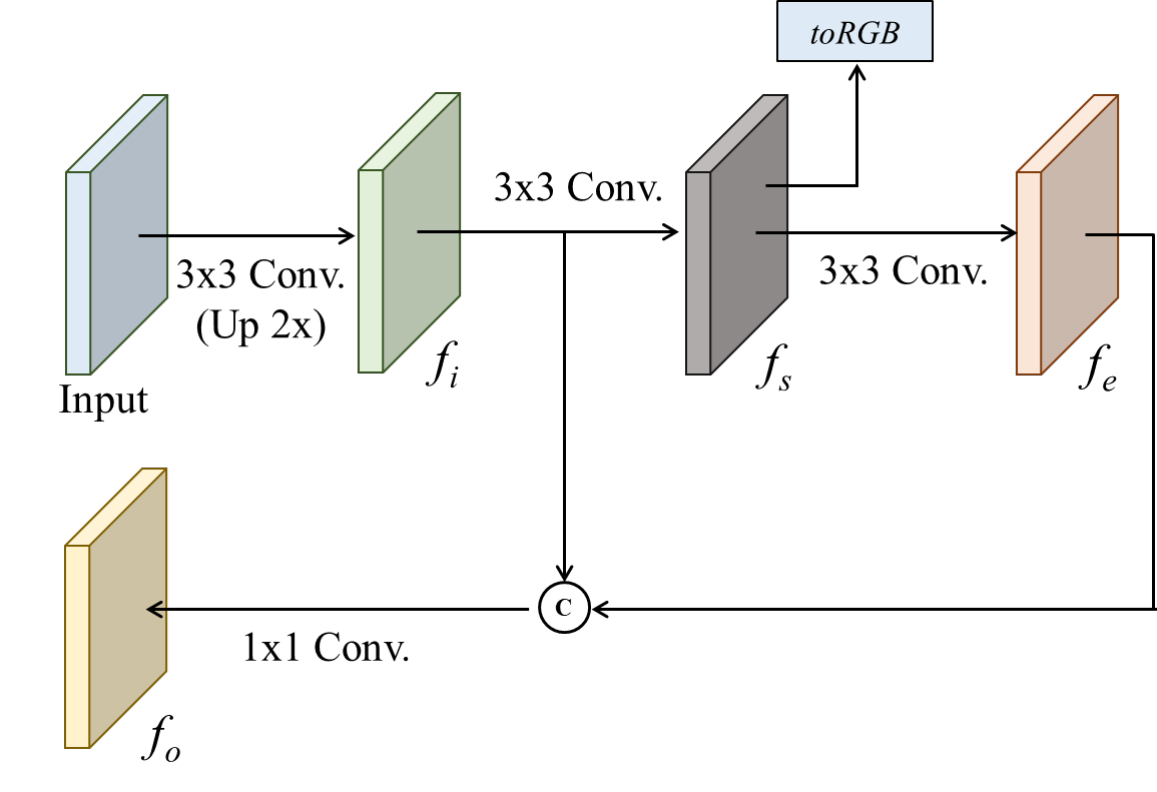}
\caption{The overall architecture of the proposed method.}
\label{fig:fig2}
\end{figure}

\subsection{Squeeze Image Connection}
\label{sec3.2}
To alleviate this problem, we propose a method that is simple but surprisingly effective. Fig.~\ref{fig:fig2} illustrates the overall framework of the proposed method, which replaces the two $3\times 3$ convolution layers present in each resolution block of StyleGAN2 (Two convolution layers of StyleGAN2 are described in Fig.~\ref{fig:fig1}). Specifically, we first produce the feature $f_i \in \mathbb{R}^{c \times h\times w}$ by using $3\times3$ convolution layer with an up-sampling technique. Then, $f_i \in \mathbb{R}^{c \times h\times w}$ is transformed into a feature $f_s \in \mathbb{R}^{c/r \times h\times w}$, which has a lower channel dimension compared to a $f_i$, through a 3x3 convolution operation (\ie, squeeze process), where \textit{r} is a user parameter that represents the channel-squeeze ratio. Here, $f_s$ is used to generate an intermediate image via the \textit{toRGB} layer; this process means reducing the dimension of the features that are concatenated to generate the output image (recall Eq.~\ref{eq8}). Subsequently, through another 3x3 convolution operation, we produce a feature $f_e\in \mathbb{R}^{c \times h\times w}$, which has the same channel dimension as $f_i$ (\ie excitation process). We agree that there might be an information loss owing to the squeeze and excitation process. To mitigate this issue, we produce the output feature $f_o\in \mathbb{R}^{c \times h\times w}$ by first concatenating $f_i$ and $f_e$ and then applying a 1x1 convolution operation. The effectiveness of this process will be proved in Section~\ref{subsec:4.4}. Note that all convolution operations, except the \textit{toRGB} layer, incorporate both modulation and demodulation processes.

These simple structural changes also help in reducing the number of network parameters. In the case of the original StyleGAN2, each resolution requires $18c^2$ convolution weight parameters, whereas the proposed method needs $(10 + 18/r)c^2$ parameters (assuming the number of input and output channels is \textit{c}, and excluding the number of \textit{Style} parameters). Thus, if the user parameter \textit{r} is greater than 2.25, it is possible to reduce the number of network variables, leading to improvements in memory efficiency. For instance, setting \textit{r} to 8 can reduce the convolution weight parameters by approximately 32\%.
Furthermore, our framework is straightforward and can be easily integrated into existing StyleGAN2-based models. In the following section, we demonstrate the effectiveness of our method in unconditional image generation across various datasets.

\begin{table}[t]
\caption{Hyperparameters used in our experiments. LR indicates learning rate. }
\begin{center}
\begin{tabular}{l | c c}
\hline
\multirow{3}*{Parameter}  & FFHQ & \multirow{2}*{CIFAR-10} \\
 & LSUN & \\
 & AFHQ & \\
 \hline
Resolution & $256\times256$ & $32\times32$ \\
Training length &  25M & 100M \\
Minibatch size & 64 & 64 \\
Minibatch stddev & 8 & 32 \\
Dataset x-flips & \checkmark / - / \checkmark & \checkmark \\
ADA~\cite{karras2020training}&  - / - / \checkmark & \checkmark \\
\hline
LR ($\times 10^{-3}$) & 2.5 & 2.5 \\
$R_1$ Reg. $\gamma$  &  1 & 0.01\\
G moving Average &  20k & 500k \\
\hline 
Mapping depth  &  8 & 2 \\
Style mixing Regularization &  \checkmark & \checkmark \\
Path length Regularization &  \checkmark & \checkmark \\
Resnet D &  \checkmark & \checkmark \\
\hline

\end{tabular}
\end{center}
\label{table:table1}
\end{table}

\section{Experiments}
\label{sec4}

\subsection{Implementation Details}
\label{subsec:4.1}
We assess the effectiveness of the proposed method across multiple datasets, including CIFAR-10~\cite{krizhevsky2009learning}, FFHQ~\cite{karras2019style}, multiple LSUN~\cite{yu15lsun} categories such as cat, horse, and church, and AFHQ~\cite{choi2020stargan}. CIFAR-10 is comprised of 50,000 small color images distributed across 10 classes, while FFHQ encompasses 70,000 high-quality facial images. AFHQ contains approximately 5,000 images per cat, dog, and wild animal face. The LSUN church, horse, and cat datasets depict respective scenes, and we have used 200,000 images per each dataset~\cite{lee2022generator}. Consistent with prior research, we have implemented horizontal flips for the CIFAR-10, FFHQ, and AFHQ datasets. All images were resized to $256 \times 256$, except CIFAR-10 images, which were maintained at $32 \times 32$. The proposed method has a single user parameter \textit{r} that determines the channel squeeze ratio. In our experiments, we empirically set the \textit{r} value as 8, except for the ablation studies. 

To prove the superior performance and generalizability of the proposed method, we executed comparative experiments against various baseline models, including StyleGAN2~\cite{karras2020analyzing}, ADA~\cite{karras2020training}, GGDR~\cite{lee2022generator}, and GLeaD~\cite{bai2023glead}. StyleGAN2 is extensively recognized in this domain for its robust performance and stable generator training. Both GGDR and GLeaD are evolved iterations of StyleGAN2, highlighting significant performance improvements in recent years. ADA applies data augmentation strategies and is particularly effective in scenarios involving datasets with limited numbers of images such as CIFAR-10 and AFHQ.

In our experiments, we systematically substituted the generator block with our proposed method across all baseline models. We maintained consistency in hyperparameters, such as learning rate and regularization weights, following the previous papers. We conducted the experiments using the official source code of each baseline models\footnote{https://github.com/NVlabs/stylegan2-ada-pytorch}$^,$ \footnote{https://github.com/naver-ai/GGDR}$^,$ \footnote{https://github.com/EzioBy/glead}. Detailed parameter settings can be found in Table~\ref{table:table1}.

\begin{table}[t]
\caption{FID and IS scores of ours and comparison methods on the CIAFR-10 datasets. We brought the numbers of conventional methods from~\cite{lee2022generator}}
\begin{center}
\begin{tabular}{c  c  c}

\hline
CIFAR-10 & FID & IS \\
\hline
DDPM~\cite{ho2020denoising} & $3.17 \pm 0.05$ & $9.46 \pm 0.11$ \\
NCSN++~\cite{song2020score} & 2.2 & 9.89 \\
WDiff~\cite{phung2023wavelet} & 4.01 & - \\
\hline
ProGAN~\cite{karras2017progressive} & 15.52 & $8.56 \pm 0.06$ \\
AutoGAN~\cite{gong2019autogan} & 12.42 & $8.55 \pm 0.10$ \\
StyleGAN2~\cite{karras2020analyzing} & $8.32 \pm 0.09$ & $9.21 \pm 0.09$\\
FSMR~\cite{kim2022feature} & 2.90 & 9.68\\
ADA~\cite{karras2020analyzing} & $2.92 \pm 0.05$ & $9.83 \pm 0.04$\\
GGDR~\cite{lee2022generator} & $2.15 \pm 0.02$ & $10.02 \pm 0.06$\\
\hline
ADA~\cite{karras2020analyzing} + Ours & $\textbf{2.50}$ & $\textbf{9.94} $\\
GGDR~\cite{lee2022generator} + Ours & $\textbf{1.98} $ & $\textbf{10.13}$ \\

\hline
\end{tabular}
\end{center}
\label{table1}
\end{table}

\subsection{Evaluation Metric}
\label{subsec:4.2}
In this study, we utilized widely recognized metrics, namely the Fréchet Inception Distance (FID)~\cite{heusel2017gans} and Precision/Recall~\cite{kynkaanniemi2019improved, sajjadi2018assessing}, to evaluate the realism of the generated images. The FID assesses the Wasserstein-2 distance between the feature distributions of real and generated images, derived from the pre-trained Inception model~\cite{szegedy2016rethinking}. FID is defined as follows: 
\begin{equation}
    \textrm{F}(p,q) = \| \mu_p - \mu_q \|_2^2 + \mathrm{trace}(C_p +C_q - 2(C_p C_q)^{\frac{1}{2}}),
\end{equation}
where $ \{\mu_p,C_p \}$ and $\{\mu_q,C_q \}$ are the mean and covariance of the samples with distributions of real and generated images, respectively. A lower FID score indicates that the generated images have better quality. Precision measures the proportion of generated images resembling those in the training set, while Recall indicates the proportion of training images that the model can replicate. For CIFAR-10, we also use Inception Score (IS)~\cite{salimans2016improved} following the previous papers~\cite{lee2022generator, karras2020training}. IS can be expressed as 
\begin{equation}
I = \mathrm{exp}(E[D_{KL}(p(l|X)||p(l))]),
\end{equation}
where \textit{l} is the class label predicted by the Inception model~\cite{szegedy2016rethinking} trained by using the ImageNet dataset~\cite{deng2009imagenet}, and $p(l|X)$ and $p(l)$ represent the conditional class distributions and marginal class distributions, respectively. Unlike the FID, the higher the IS score, the better the quality of the generated image. For our analysis, we generated 50,000 samples and calculated these metrics accordingly. 

\begin{figure}
\centering
\includegraphics[width=0.8\linewidth]{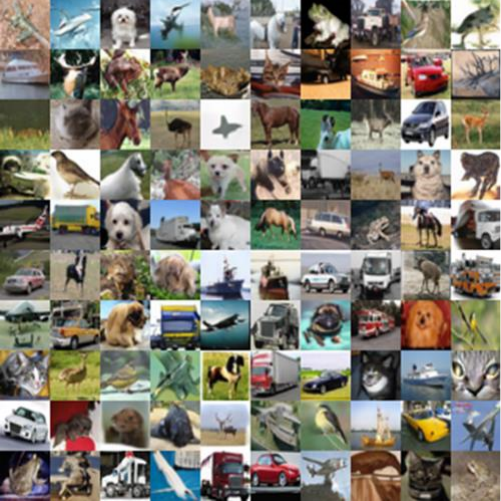}
\caption{Sample images of the proposed method on the CIFAR-10 dataset.}
\label{fig:fig3}
\end{figure}

\begin{table*}[t]
\caption{Comparison on FFHQ, LSUN Church, LSUN Horse, and LSUN Cat. P and R denote precision and recall, respectively. Lower FID and higher precision and recall mean better performance. The bold numbers indicate the best FID, P, and R for each dataset. We brought the numbers of conventional methods from~\cite{lee2022generator, bai2023glead}}
\begin{center}
\begin{tabular}{r c c c c c c c c c c c c c c c c }
\hline
\\[-1em]
\multirow{2}*{Method} & & \multicolumn{3}{c}{FFHQ} & & \multicolumn{3}{c}{LSUN Church} &  & \multicolumn{3}{c}{LSUN Horse} & & \multicolumn{3}{c}{LSUN Cat} \\
\\[-1em] \cline{3-5} \cline{7-9} \cline{11-13} \cline{15-17}  \\[-1em]

&  & FID$\downarrow$ & P$\uparrow$ & R$\uparrow$ & & FID$\downarrow$ & P$\uparrow$ & R$\uparrow$ & & FID$\downarrow$ & P$\uparrow$ & R$\uparrow$ & & FID$\downarrow$ & P$\uparrow$ & R$\uparrow$ \\
\\[-1em] \hline \\[-1em]

UT~\cite{bond2022unleashing} & & 6.11 & \textbf{0.73} & 0.48 & & 4.07 & \textbf{0.71} & 0.45 & & - & - & - & & - & - & -\\
Polarity~\cite{humayun2022polarity} & & - & - & - & & 3.92 & 0.61 & 0.39 & & - & - & - & & 6.39 & \textbf{0.64} & 0.32\\
\hline \\[-1em]

StyleGAN2~\cite{karras2020analyzing} & & 3.71 & 0.69 & 0.44 & & 3.97 & 0.59 & 0.39 & & 3.62 & 0.63 & 0.36 & & 7.98 & 0.60 & 0.27 \\
StyleGAN2~\cite{karras2020analyzing} + Ours & & 3.37 & 0.69 & 0.48 & & 3.86 & 0.59 & 0.46 & & 2.95 & 0.62 & 0.42 & & 7.28 & 0.59 & 0.30 \\
\\[-1em] \hline \\[-1em]

GGDR~\cite{lee2022generator} & & 3.14 & 0.69 & 0.50 & & 3.15 & 0.61 & 0.46 & & 2.50 & \textbf{0.64} & 0.43 & & 5.28 & 0.58 & 0.38 \\
GGDR~\cite{lee2022generator} + Ours & & 2.90 & 0.68 & 0.52 & & 2.61 & 0.59 & 0.52 & & 2.45 & 0.60 & \textbf{0.48} & &\textbf{5.03} & 0.58 & 0.39  \\
\\[-1em] \hline  \\[-1em]

GLeaD~\cite{bai2023glead} & & 3.24 & 0.69 & 0.47 & & 2.82 & 0.62 & 0.43 & & - & - & - &  & - & - & -\\
GLeaD~\cite{bai2023glead}+Ours & & \textbf{2.69} & 0.68 & \textbf{0.53} & & \textbf{2.29} & 0.59 & \textbf{0.53} & & \textbf{2.41} & 0.61 & \textbf{0.48} & & 5.43 & 0.56 & \textbf{0.40} \\
\\[-1em]

\hline
\end{tabular}
\end{center}
\label{table2}
\end{table*}

\begin{figure*}
\centering
\includegraphics[width=0.9\linewidth]{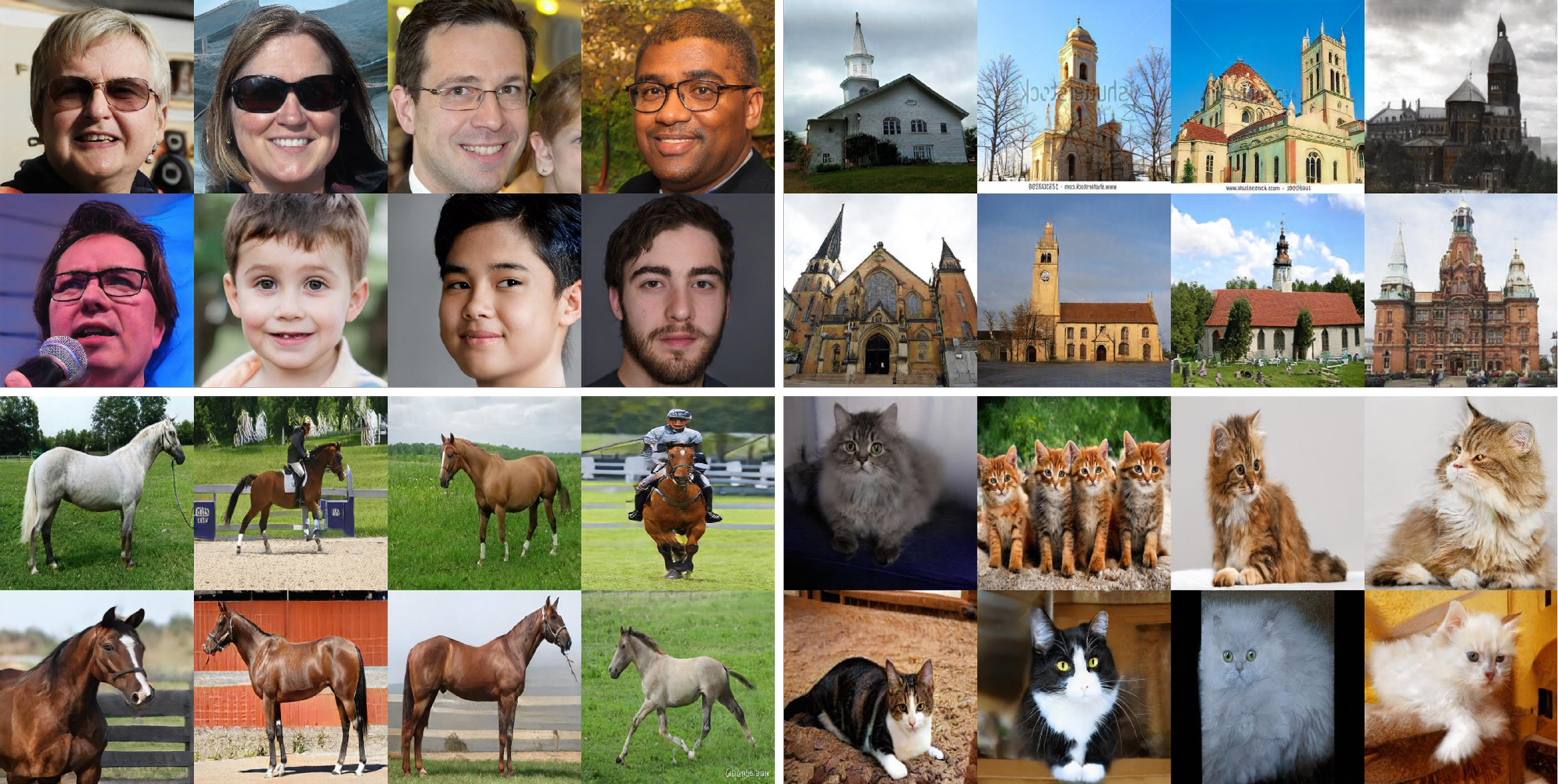}
\caption{Selective samples generated by our method. For FFHQ and LSUN datasets, we show the results of GGDR with our method.}
\label{fig_FFHQ}
\end{figure*}

\subsection{Experimental Results}
\label{subsec:4.3}
To evaluate the performance enhancement brought by our method, we first conducted experiments on the CIFAR-10 dataset. We compare the performance of our method not only with the leading GANs-based models~\cite{karras2017progressive, gong2019autogan, karras2020analyzing, kim2022feature, karras2020analyzing, lee2022generator} but also with the latest diffusion-based models~\cite{ho2020denoising, song2020score, phung2023wavelet}. Table~\ref{table1} shows the FID and IS scores of ours and comparison methods on the CIFAR-10 dataset. Specifically, we incorporated our method into ADA~\cite{karras2020training}, recognized as a benchmark model for unconditional image generation, particularly effective on small datasets. As shown in Table~\ref{table1}, the proposed method successfully improves the ADA performance in terms of both FID and IS scores. Moreover, we applied our proposed method to GGDR~\cite{lee2022generator} which already outperforms ADA. As demonstrated by our experimental results, the proposed method not only significantly boosts the GGDR performance but also achieves state-of-the-art performance. Fig.~\ref{fig:fig3} shows the samples of our method on the CIFAR-10 dataset. 

To prove the enhanced capability of the proposed method in generating high-resolution images, we conducted experiments using multiple datasets such as FFHQ, LSUN-Church, LSUN-Horse, and LSUN-Cat. We set the StyleGAN2, GGDR, and GLeaD~\cite{bai2023glead} methods as our baseline models following the previous papers~\cite{lee2022generator, bai2023glead}. Table~\ref{table2} presents the FID and Precision/Recall scores for both the proposed method and the comparative methods. Specifically, we first assess the performance enhancement brought by the proposed method, using StyleGAN2 as the baseline model. As shown in Table~\ref{table2}, the proposed method not only achieves superior performance across all datasets used in the experiments compared to prior works such as UT~\cite{bond2022unleashing} and Polarity~\cite{humayun2022polarity} but also marginally enhances the performance of StyleGAN. In particular, the proposed method improves the Recall with the large gap compared to StyleGAN2, which indicates that the proposed method generates more diverse images and is less prone to mode collapse. Moreover, we proceeded with experiments using GGDR and GLeaD, which enhance the performance of StyleGAN2, as our baseline models. As depicted in Table~\ref{table2}, the proposed method improves performance across all datasets, and notably, in the case of the GLeaD baseline, both FID and Recall metrics exhibit significant enhancements. Therefore, these experimental results confirm that the proposed method consistently boosts performance, regardless of the baseline. The samples of the generated image are presented in Fig.~\ref{fig_FFHQ}. As shown in Fig.~\ref{fig_FFHQ}, the proposed method produces high-quality synthetic images that appear remarkably real.

\begin{figure}
\centering
\includegraphics[width=0.9\linewidth]{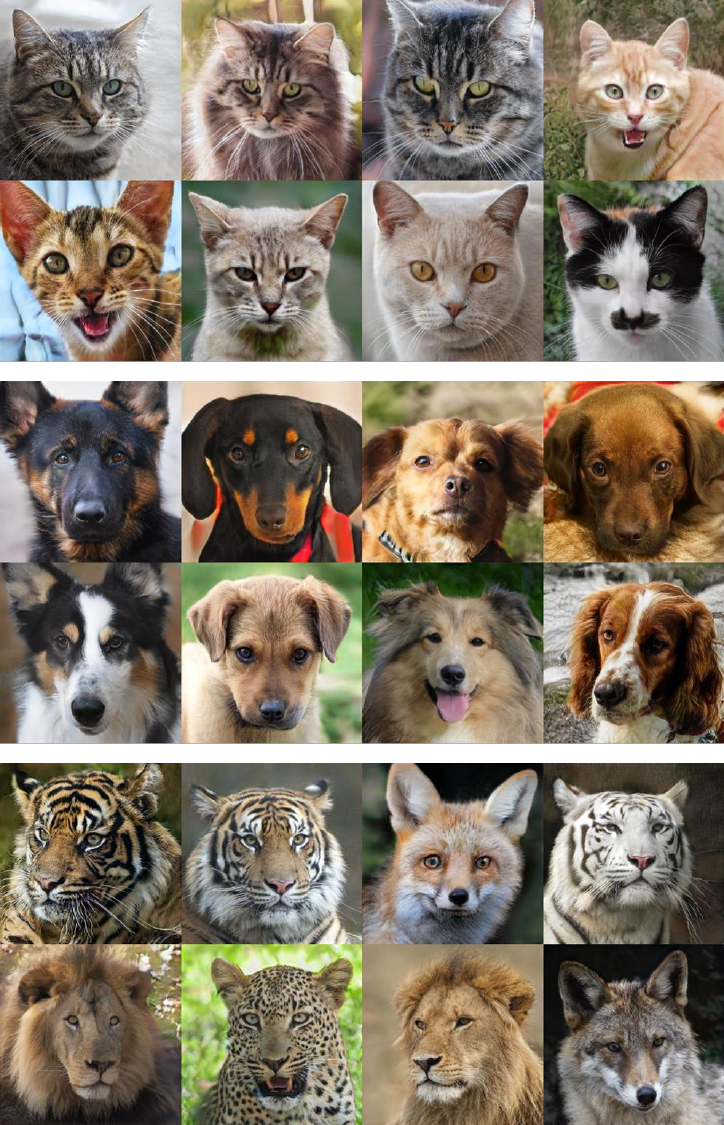}
\caption{Selective samples generated by our method. For AFHQ datasets, we show the results of ADA with our method.}
\label{fig5}
\end{figure}

To show the robustness of the proposed method against the small dataset, we conducted experiments using the AFHQ dataset~\cite{choi2020stargan} by setting the ADA~\cite{karras2020training} as the baseline model. Table~\ref{table2_ada} presents the experimental results. Since the existing paper~\cite{kim2022feature} only compared FID scores, this brief also focuses solely on FID comparisons. As shown in Table~\ref{table2_ada}, the proposed method improves the baseline in terms of FID scores with a large gap; the proposed method enhances the performance of the baseline model successfully, even on datasets with a smaller number of images. These results indicate that the proposed method is not only effective but also highly adaptable to various types of data, which in turn, enhances its utility and relevance in the various applications. Fig.~\ref{fig5} shows samples of our method on the AFHQ datasets.

\subsection{Ablation Studies}
\label{subsec:4.4}
The proposed method includes a user-defined variable \textit{r} that determines the degree of channel compression. While compressing the channels can reduce the number of network parameters, it may also decrease performance. Thus, making the selection of an appropriate \textit{r} value is crucial. To determine the optimal value, we conducted ablation studies on the FFHQ and LSUN church datasets. We use the StyleGAN2 as the baseline model. As summarized in Table~\ref{table3}, we can observe that as the \textit{r} value increases, the number of parameters in G decreases. However, there is a corresponding decline in performance in terms of FID metric as the number of G parameters decreases. For instance, at $r=16$, the performance on the LSUN church dataset shows weak performance compared to the baseline model. Conversely, $r=4$ and $r=8$ not only show superior performance compared to the baseline model but also the added benefit of reduced parameter count. In terms of the Recall metric, the proposed method consistently outperforms the baseline model, irrespective of the \textit{r} value. We chose $r=8$ as it significantly reduces the parameters of the network while still maintaining respectable performance.

\begin{table}[t]
\caption{FID Scores on AFHQ Dog, Cat and Wild. Lower FID means better performance. The bold numbers indicate the best FID for each dataset. We brought the numbers of conventional methods from~\cite{kim2022feature}}
\begin{center}
\begin{tabular}{r c c c c }
\hline
\\[-1em]
Method & & Dog & Cat & Wild \\
\\[-1em] \hline \\[-1em]

StyleGAN2~\cite{karras2020analyzing} & & 19.65 & 8.37 & 4.17\\
DiffAug~\cite{zhao2020differentiable} & & 16.92 & 6.39 & 4.39 \\
\\[-1em] \hline \\[-1em]

ADA~\cite{karras2020training} & & 13.56 & 6.64 & 3.74 \\
ADA~\cite{karras2020training} + FSMR~\cite{kim2022feature} & & 11.76 & 5.71 & 3.24 \\
ADA~\cite{karras2020training} + Ours & & \textbf{9.68} &\textbf{4.38} & \textbf{2.63} \\

\\[-1em]

\hline
\end{tabular}
\end{center}
\label{table2_ada}
\end{table}

\begin{figure}
\centering
\includegraphics[width=0.9\linewidth]{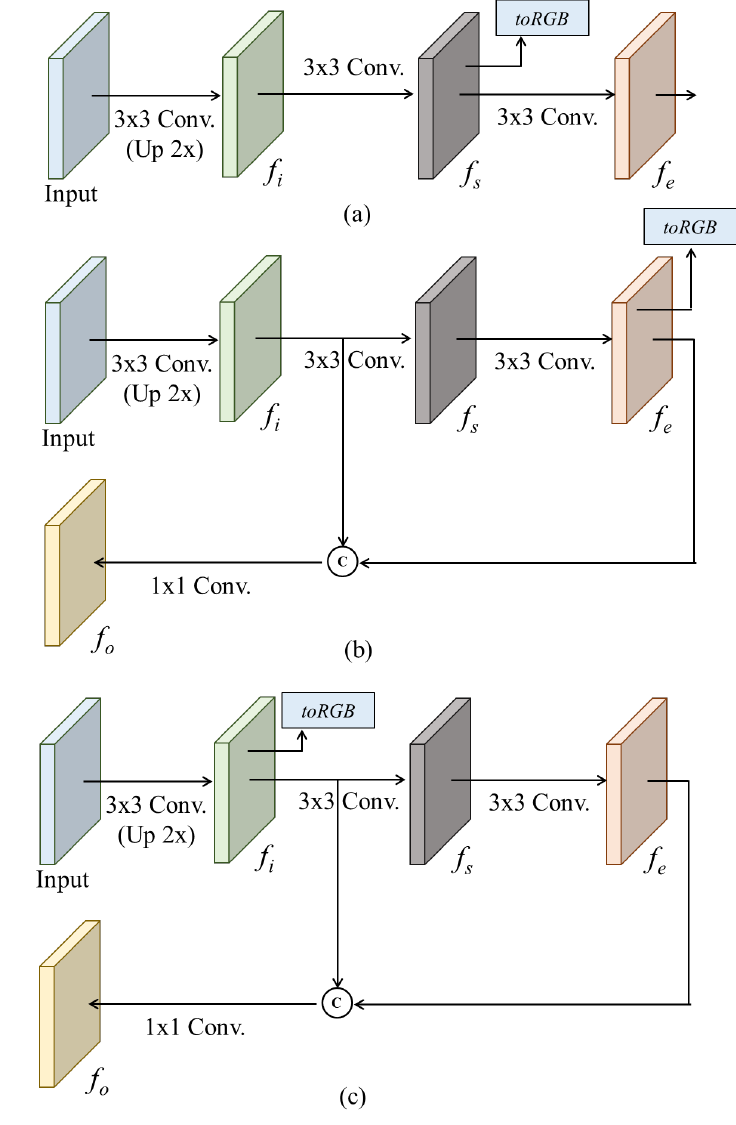}
\caption{The modified architectures of the proposed method utilized for the ablation studies.}
\label{fig:fig4}
\end{figure}

As mentioned in Section~\ref{sec3.2}, there is a need to validate the necessity of the feature blending process, which involves the concatenation of features followed by a $1\times1$ convolution. To this end, we build a modified structure without the feature blending process (as shown in Fig.~\ref{fig:fig4}(a)) and evaluate its performance. Table~\ref{table4} summarizes the experimental results. The method without the feature blending process shows lower performance compared to the proposed method, yet it still exhibits comparable or superior performance relative to the baseline model. These results indicate that the significant contribution to performance improvement stems from the image squeeze connection rather than the feature blending process. We believe that these results indirectly highlight the limitations of the existing image skip connection method and the superiority of our approach.

\begin{table}[t]
\caption{Ablation studies about the \textit{r} value on FFHQ and LSUN Church datasets.}
\begin{center}
\begin{tabular}{l c c c c c c c }
\hline
\\[-1em]
\multirow{2}*{Method} & \multirow{2}*{G Parmas.}  &  & \multicolumn{2}{c}{FFHQ} & & \multicolumn{2}{c}{LSUN Church} \\
\\[-1em] \cline{4-5} \cline{7-8} \\[-1em]

& &  & FID$\downarrow$ & R$\uparrow$ & & FID$\downarrow$ & R$\uparrow$  \\
\\[-1em] \hline \\[-1em]

StlyeGAN2 & 24.80M & & 3.71 & 0.44 & & 3.97 & 0.39  \\
\hline \\[-1em]
+ Ours ($r = 4$)  & 24.25M & & 3.29 & 0.49 & & 3.71 & 0.46  \\
+ Ours ($r = 8$)  & 21.80M & & 3.37 & 0.48 & & 3.86 & 0.46  \\
+ Ours ($r = 16$)  & 20.56M & & 3.44 & 0.49 & & 4.25 & 0.45  \\

\hline
\end{tabular}
\end{center}
\label{table3}
\end{table}

Furthermore, we also measured the performance variations according to the placement of the \textit{toRGB} layer (Figs.~\ref{fig:fig4}(b) and ~\ref{fig:fig4}(c)). Specifically, Fig~\ref{fig:fig4}(b) illustrates the use of the \textit{toRGB} layer before the squeeze process, while Fig.~\ref{fig:fig4}(c) depicts its application after the excitation process. It is important to note that both methods employ features that have high-dimension to the \textit{toRGB} layer. In other words, similar to the traditional image skip connection, high-dimensional intermediate features are concatenated through the \textit{toRGB} layer. As shown in Table~\ref{table6}, it can be observed that both cases exhibit lower performance compared to the proposed method. Additionally, they also show decreased performance compared to the baseline model. These results indicate that the performance improvement in the proposed method is not merely due to the presence of squeeze and excitation processes. Instead, it suggests that utilizing low-dimensional features to the \textit{toRGB} layer enhances performance. Based on these experimental results, we consider the proposed method as a solution that addresses the issues with the traditional image skip connection and significantly aids in improving the performance of StyleGAN2-based models.

\section{Conclusion}
\label{sec5}
In this brief, we identify and mathematically analyze the issue with the image skip connection in StyleGAN2, detailing the reasons behind the problem. Inspired by our observations, we propose a new skip connection method, called image squeeze connection, aimed at enhancing the image synthesis performance. The proposed technique not only shows superior performance compared to strong baseline models but also reduces the number of network parameters. Furthermore, due to its simplicity, the proposed method can be easily applied to existing baseline models. To prove the superiority of the proposed method through comprehensive experiments conducted on various baseline models developed using StyleGAN2. Furthermore, due to the excellence of StyleGAN2, many models based on it continue to be developed recently. Therefore, we anticipate that our proposed method could be widely adopted in the relevant fields.

\begin{table}[t]
\caption{Ablation studies about the feature blending process on FFHQ and LSUN Church datasets. FBP indicates the feature blending process. The bold numbers indicate the best FID and Recall}
\begin{center}
\begin{tabular}{l c c c c c c c }
\hline
\\[-1em]
\multirow{2}*{Method} & \multirow{2}*{G Parmas.}  &  & \multicolumn{2}{c}{FFHQ} & & \multicolumn{2}{c}{LSUN Church} \\
\\[-1em] \cline{4-5} \cline{7-8} \\[-1em]

& &  & FID$\downarrow$ & R$\uparrow$ & & FID$\downarrow$ & R$\uparrow$  \\
\\[-1em] \hline \\[-1em]

StlyeGAN2 & 24.80M & & 3.71 & 0.44 & & 3.97 & 0.39  \\
\hline \\[-1em]
+ Ours   & 21.80M & & \textbf{3.37} & \textbf{0.48} & & \textbf{3.86} & \textbf{0.46}\\
+ Ours w/o FBP  & 21.75M & & 3.40 & 0.46 & & 4.05 & 0.43  \\

\hline
\end{tabular}
\end{center}
\label{table4}
\end{table}

\begin{table}[t]
\caption{Ablation studies about the location of \textit{toRGB} layer on FFHQ and LSUN Church datasets. The bold numbers indicate the best FID and Recall}
\begin{center}
\begin{tabular}{l c c c c c c }
\hline
\\[-1em]
\multirow{2}*{Method}  &  & \multicolumn{2}{c}{FFHQ} & & \multicolumn{2}{c}{LSUN Church} \\
\\[-1em] \cline{3-4} \cline{6-7} \\[-1em]

&  & FID$\downarrow$ & R$\uparrow$ & & FID$\downarrow$ & R$\uparrow$  \\
\\[-1em] \hline \\[-1em]

StlyeGAN2  & & 3.71 & 0.44 & & 3.97 & 0.39  \\
\hline \\[-1em]
+ Ours    & & \textbf{3.37} & \textbf{0.48} & & \textbf{3.86}& \textbf{0.46}  \\
+ Ablation1 (Fig.~\ref{fig:fig4}(b))  & & 3.91 & 0.48 & & 4.11 & 0.45  \\
+ Ablation2 (Fig.~\ref{fig:fig4}(c))  & & 3.48 & 0.48 & & 4.20 & 0.42  \\

\hline
\end{tabular}
\end{center}
\label{table6}
\end{table}


\bibliographystyle{IEEEtran}
\bibliography{egbib.bib}

%



\end{document}